%% file: anonymous-submission-latex-2025.tex
\documentclass[letterpaper]{article} 
\usepackage{aaai25}  
\usepackage{times}  
\usepackage{helvet}  
\usepackage{courier}  
\usepackage[hyphens]{url}  
\usepackage{graphicx} 
\usepackage{natbib}  
\usepackage{caption} 
\usepackage{algorithm}
\usepackage{algorithmic}

\usepackage{amssymb}
\usepackage{newfloat}
\usepackage{listings}
\DeclareCaptionStyle{ruled}{labelfont=normalfont,labelsep=colon,strut=off} 

\usepackage{multirow}

\usepackage{booktabs}
\usepackage{courier}
\usepackage{listings}
\usepackage{graphicx}
\usepackage{caption}
\usepackage{array}
\usepackage{float}

\usepackage{siunitx}

\usepackage[switch]{lineno}

\usepackage{url}

\lstdefinestyle{customstyle}{
    basicstyle=\ttfamily\footnotesize,
    commentstyle=\color{gray},
    keywordstyle=\color{blue},
    stringstyle=\color{red},
    numbers=none,    
    tabsize=2,
    extendedchars=true,
    backgroundcolor=\color{white},
    showspaces=false,
    showtabs=false,
    frame=single,
    rulecolor=\color{black},
    captionpos=b,
    breaklines=true,
    breakatwhitespace=false,
    escapeinside={\%*}{*)},
    keepspaces=true,
}

\floatstyle{ruled}
\newfloat{listing}{tb}{lst}{}
\floatname{listing}{Listing}
\title{ReXplain: Translating Radiology into Patient-Friendly Video Reports}
\author{
{Luyang Luo}$^{1}$, {Jenanan Vairavamurthy}$^2$, {Xiaoman Zhang}$^1$, \\ {Abhinav Kumar}$^3$, {Ramon R. Ter-Oganesyan}$^4$, {Stuart T. Schroff}$^4$, \\{Dan Shilo}$^2$, {Rydhwana Hossain}$^5$, {Mike Moritz}$^6$, {Pranav Rajpurkar}$^1$ 
    }

\affiliations{
    $^1$Department of Biomedical Informatics, Harvard University, USA \\
$^2$Department of Radiology, Icahn School of Medicine, Mount Sinai, USA \\
$^3$Department of Computer Science, Stanford University, USA \\
$^4$Department of Radiology, Los Angeles General Medical Center, USA\\
$^5$Department of Radiology, University of Maryland, USA\\
$^6$SSM Health, Saint Louis University, USA
}

\usepackage{bibentry}

\begin{document}

\maketitle

\begin{abstract}
Radiology reports, designed for efficient communication between medical experts, often remain incomprehensible to patients.
This inaccessibility could potentially lead to anxiety, decreased engagement in treatment decisions, and poorer health outcomes, undermining patient-centered care.
We present ReXplain (Radiology eXplanation), an innovative AI-driven system that translates radiology findings into patient-friendly video reports. 
ReXplain uniquely integrates a large language model for medical text simplification and text-anatomy association, an image segmentation model for anatomical region identification, and an avatar generation tool for engaging interface visualization.
ReXplain enables producing comprehensive explanations with plain language, highlighted imagery, and 3D organ renderings in the form of video reports.
To evaluate the utility of ReXplain-generated explanations, we conducted two rounds of user feedback collection from six board-certified radiologists.
The results of this proof-of-concept study indicates that ReXplain could accurately deliver radiological information and effectively simulate one-on-one consultation, shedding light on enhancing patient-centered radiology with potential clinical usage.
This work demonstrates a new paradigm in AI-assisted medical communication, potentially improving patient engagement and satisfaction in radiology care, and opens new avenues for research in multimodal medical communication.
\end{abstract}

%

\input{content/01-intro}

\input{content/02-relatedwork}

\input{content/02-design}

\input{content/03-output}

\input{content/04-user}

\input{content/05-discussion}

\bibliography{aaai25}

\onecolumn

\appendix
\section{Appendix}\label{apd:first}

\subsection{Prompt for matching report phrases to organs.}

\begin{lstlisting}[style=customstyle]
Act as an experienced clinician with radiology knowledge, please extract a list of the entities (organs) with positive findings out of the given text report. If the detected organ belongs to the following list, then add this exact organ into the output list; otherwise look into the relationships between the detected organ and those in the list, find the closest one (in terms of belonging relationship) in the list and put that one into the output list. The output should strictly formatted a list of organs, no other text should be provided.

***
organs = [`lung', 
          `left lung', 
          `left lung lower lobe', 
          `left lung upper lobe', ... 
          `larynx supraglottis', 
          `muscle']
***

Here is an example to show that only organs with positive/abnormal findings will be extracted, and organs described withouth abnormal findings would not be shown:

- User: Thickening of the bronchial wall and peribronchial budding tree-like reticulonodular densities are observed in the bilateral lower lobes. Other mediastinal main vascular structures, heart contour, size are normal. 

- Output: left lung lower lobe,right lung lower lobe

*** 

If no abnormalities were found, return none.
\end{lstlisting}

\subsection{Prompt for report phrase explanation.}
\begin{lstlisting}[style=customstyle]
You are an experienced clinician with radiology knowledge. Given the description of a patient's CT imaging report, please help explain it to the patient in a friendly, easy-to-understand, and brief way. Your explanation should contain 1. A brief concise description in layman language of what this abnormality means; including what it indicates clinically; 2. How it should looks like in the CT image; and 3. How a normal person's CT would be like without the abnormality. Generate it in a smooth way as if you are guiding the patient to look at the two different CT images. Your output should be in the format of a json file. Here is an example:

\noindent***

User: Left subclavian vein collapsed.

Output: {"abnormality_explanation": ``Your left subclavian vein is found to be collapsed. This vein is an important blood vessel that carries blood from your arm back to your heart. When we say it's collapsed, it means that the vein is not as open as it should be, which might affect how blood flows through it.",

"input_scan_appearance": "If we look at your CT image, the left subclavian vein will appear much narrower or even almost closed in some areas. It might look like a thin line instead of a round, hollow tube.",

"normal_scan_appearance": "In a normal CT image, this vein would appear open and round, kind of like a small, hollow tube. It would look uniform and not pinched or narrow."}

***
\end{lstlisting}

\newpage

\subsection{List of Organs.}
\begin{lstlisting}[style=customstyle]
organs = ['lung', 'left lung', 'left lung lower lobe', 'left lung upper lobe', 'right lung',
'right lung lower lobe', 'right lung middle lobe', 'right lung upper lobe', 
'lung lower lobe','lung upper lobe', 'trachea and bronchie', 'trachea', 'bronchie', 
'mediastinum','brachiocephalic trunk', 'brachiocephalic vein', 'left brachiocephalic vein', 
'right brachiocephalic vein', 'superior vena cava', 'aorta', 'pulmonary artery', 'thymus',
'pulmonary vein', 'mediastinal tissue', 'subclavian artery', 'left subclavian artery',
'right subclavian artery', 'heart', 'heart atrium', 'left heart atrium', 
'right heart atrium', 'heart ventricle', 'left heart ventricle', 'right heart ventricle', 
'heart ascending aorta', 'heart tissue', 'myocardium', 'left auricle of heart',
'esophagus', 'cervical esophagus', 'cricopharyngeal inlet', 'pleura', 'bone',
'spinal cord', 'spinal canal', 'vertebrae', 'cervical vertebrae',
'cervical vertebrae 1 (c1)', 'cervical vertebrae 2 (c2)', 'cervical vertebrae 3 (c3)', 
'cervical vertebrae 4 (c4)', 'cervical vertebrae 5 (c5)', 'cervical vertebrae 6 (c6)',
'cervical vertebrae 7 (c7)', 'thoracic vertebrae', 'thoracic vertebrae 1 (t1)',
'thoracic vertebrae 2 (t2)', 'thoracic vertebrae 3 (t3)', 'thoracic vertebrae 4 (t4)', 
'thoracic vertebrae 5 (t5)', 'thoracic vertebrae 6 (t6)','thoracic vertebrae 7 (t7)', 
'thoracic vertebrae 8 (t8)', 'thoracic vertebrae 9 (t9)','thoracic vertebrae 10 (t10)',
'thoracic vertebrae 11 (t11)','thoracic vertebrae 12 (t12)','lumbar vertebrae', 
'lumbar vertebrae 1 (l1)', 'lumbar vertebrae 2 (l2)','lumbar vertebrae 3 (l3)',
'lumbar vertebrae 4 (l4)', 'lumbar vertebrae 5 (l5)', 'lumbar vertebrae 6 (l6)',
'sacral vertebrae 1 (s1)', 'clavicle', 'left clavicle', 'right clavicle', 'scapula',
'left scapula', 'right scapula', 'humerus', 'left humerus', 'right humerus', 'femur',
'left femur', 'right femur', 'head of femur', 'left head of femur', 'right head of femur',
'rib', 'left rib', 'right rib', 'rib 1', 'left rib 1', 'right rib 1', 'rib 2', 'left rib 2',
'right rib 2', 'rib 3', 'left rib 3', 'right rib 3', 'rib 4', 'left rib 4', 'right rib 4',
'rib 5', 'left rib 5', 'right rib 5', 'rib 6', 'left rib 6', 'right rib 6', 'rib 7',
'left rib 7', 'right rib 7', 'rib 8', 'left rib 8', 'right rib 8', 'rib 9', 'left rib 9', 
'right rib 9', 'rib 10', 'left rib 10', 'right rib 10', 'rib 11', 'left rib 11',
'right rib 11', 'rib 12', 'left rib 12', 'right rib 12', 'rib cartilage', 'costal cartilage',
'sternum', 'manubrium of sternum', 'eustachian tube bone', 'left eustachian tube bone',
'right eustachian tube bone', 'thyroid', 'thyroid gland', 'left thyroid', 'right thyroid',
'breast', 'left breast', 'right breast', 'abdomen', 'abdominal tissue', 'adrenal gland',
'left adrenal gland', 'right adrenal gland', 'colon', 'duodenum', 'gallbladder', 'intestine',
'small bowel', 'kidney', 'left kidney', 'right kidney', 'liver', 'left lobe of liver',
'left lateral inferior segment of liver', 'left lateral superior segment of liver',
'left medial segment of liver', 'right lobe of liver', 
'right anterior inferior segment of liver', 'right anterior superior segment of liver', 
'right posterior inferior segment of liver', 'right posterior superior segment of liver', 
'liver vessel', 'caudate lobe', 'pancreas', 'portal vein and splenic vein', 'rectum',
'renal artery', 'renal vein', 'spleen', 'stomach', 'celiac trunk', 'thoracic cavity',
'prostate', 'urinary bladder', 'carotid artery', 'common carotid artery', 
'internal carotid artery', 'left carotid artery', 'left common carotid artery',
'left internal carotid artery', 'right carotid artery', 'right common carotid artery',
'right internal carotid artery', 'iliac artery', 'iliac vena', 'iliac vein',
'left iliac artery', 'left iliac vena', 'right iliac artery', 'right iliac vena',
'inferior vena cava', 'internal jugular vein', 'larynx', 'larynx glottis',
'larynx supraglottis', 'muscle']
\end{lstlisting}

\end{document}

%% file: content/01-intro.tex
\section{Introduction}
\label{sec:intro}

\begin{figure}[t]
  \includegraphics[width=\linewidth]{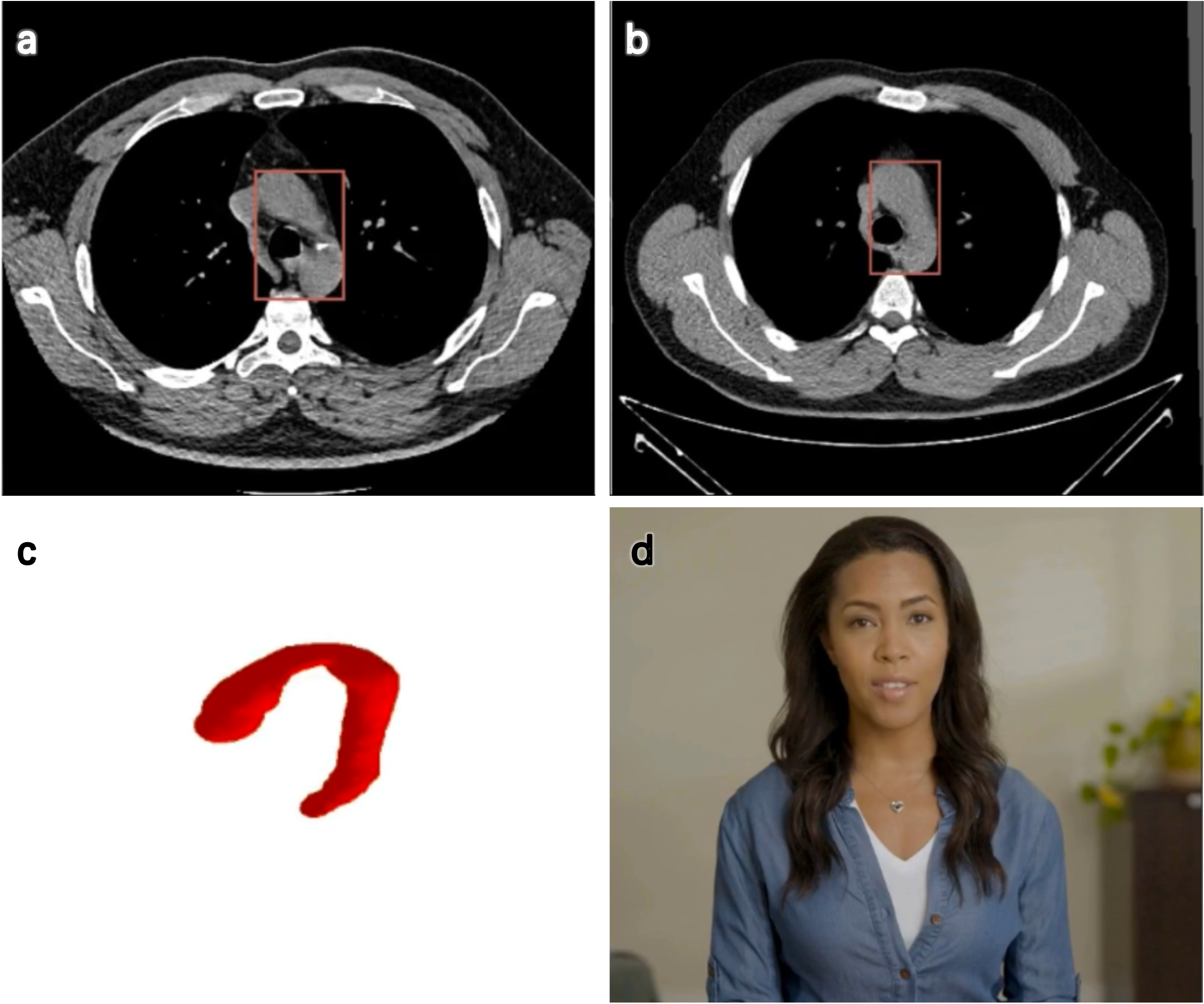}
  \caption{\textbf{Key elements of ReXplain-generated video reports.} \textbf{a}. The image of the patient with the key finding, calcification in the aorta, highlighted in a bounding box; \textbf{b}. A normal reference image registered with the query image, and the aorta is also highlighted; \textbf{c}. The reconstruction rendering of the aorta; \textbf{d}. A talking Avatar explaining the text reports with lay-language.}
\label{fig:teaser}
\end{figure}

\textbf{Background and Motivation.}
Radiology reports play a crucial role in patient care.
However, there has been a growing disparity between the demand for medical imaging services and the capacity of radiologists to maintain high reporting standards \citep{konstantinidis2024shortage}, exacerbated by a global shortage of 122,000 radiologists by 2032 \cite{heiser2019new}. This imbalance leads to radiologist burnout, delays in critical care delivery, and widening healthcare disparities \cite{bluth2022recommendation,kemp2017patient}. 
Despite a consistent, strong desire among patients to access these reports \citep{halaska2019patient,garry2020patient}, these documents often remain inaccessible to the very individuals they concern most: the patients themselves. The complex medical terminology and abbreviations, designed for efficient communication among healthcare professionals, create a significant barrier to patient understanding. This can lead to misunderstandings, anxiety, and decreased engagement in treatment decision, ultimately increasing the burden on doctors and potentially impacting health outcomes.

\textbf{Patient-centered Radiology.}
Patient-centered radiology, which involves radiologists actively engaging with patients by introducing imaging procedures, communicating results directly, and addressing concerns, has gained increasing attention \citep{itri2015patient,kemp2017patient,dutruel2023patient}. 
Recent years have also witnessed a shift towards making radiology reports and images directly available to patients, extending their utility beyond interprofessional communication \cite{halaska2019patient,garry2020patient}.
Particularly, both healthcare providers and patients have responded positively to patient-friendly reports, as these can enhance patient understanding and overall satisfaction with their care \citep{lopez2024user}.
To achieve patient-centered radiology efficiently, it is suggested to leverage digital tools to improve patient's understanding of their imaging findings, increase compliance with follow-up exams, and enable convenient image sharing \cite{dutruel2023patient}.

\textbf{Current Approaches and Limitations.}
Various approaches have been proposed to enhance patient understanding of radiology reports, ranging from supplementary medical terminology explanations to structured reporting formats \citep{cook2017patients,lourenco2020optimizing}. Among these, video reports have emerged as a particularly promising direction, as they effectively demonstrate improved patient comprehension and satisfaction by seamlessly connecting written findings with visual guidance \citep{recht2022video}. Through synchronized visual and verbal explanations, these radiologist-created video reports successfully bridge the gap between complex medical imaging and patient understanding \citep{balkman2016audio,neto2019evaluation}. Despite their effectiveness, the creation of such video reports introduces substantial additional demands on radiologists' already constrained time and resources. This burden becomes increasingly unsustainable in light of the aforementioned radiologist shortage, further exacerbating the existing workforce challenges. As such, the video duration is inherently constrained by the availability of expert clinicians' time, limiting the depth and breadth of explanations. This limitation becomes particularly pronounced for complex volumetric imaging modalities like Computed Tomography (CT) and Magnetic Resonance Imaging (MRI), where comprehensive analysis requires meticulous slice-by-slice navigation and explanation.

\textbf{AI in Radiology Communication.}
Recent advances in language processing \citep{eriksen2023use,wei2022emergent}, medical image computing \citep{zhou2021review}, and cross-modal generation \citep{croitoru2023diffusion,acosta2022multimodal} have demonstrated remarkable potential in transforming healthcare and medicine \citep{rajpurkar2022ai}. In particular, Large Language Models (LLMs) have emerged as a promising solution for bridging the communication gap between medical professionals and patients. These models excel at simplifying radiology reports while preserving factual accuracy and completeness \citep{lyu2023translating,doshi2023utilizing,jeblick2024chatgpt}. Such capabilities suggest significant potential for enhancing patient experience and engagement with their medical information \citep{elkassem2023potential}. However, current AI applications in radiology communication predominantly focus on text simplification, leaving the crucial aspect of medical image interpretation largely unexplored. The primary technical challenge involves aligning textual descriptions with visual findings on radiology images, necessitating AI models capable of advanced multimodal comprehension.

\textbf{Our Approach and Contributions.}
We propose \textbf{ReXplain}, an AI-driven radiology video report generation pipeline addressing this challenge. \textbf{The core innovation of our approach lies not in the development of new AI models, but in the novel integration of existing state-of-the-art modules to create a functional, end-to-end system.} By combining advanced language models, image segmentation techniques, and avatar generation technology, we've created a proof-of-concept system that produces comprehensive, visually-enhanced explanations of radiology findings. 

\noindent \vspace{3pt} \textbf{The key components of ReXplain include:}
\begin{itemize}
    \item A large language model (LLM) for translating complex radiology reports into plain language and connecting findings to corresponding anatomies;
    \item A segmentation model for locating and highlighting relevant anatomical regions in CT scans;
    \item An avatar generation tool for creating a virtual presenter.
\end{itemize}

\textbf{The innovation lies in how these components are orchestrated to work together seamlessly, creating a direct link between simplified textual explanations and corresponding areas of interest in medical images.}
Our system goes beyond mere text simplification or image analysis. It produces video reports that mimic the nuanced explanations typically provided by radiologists to patients, complete with image display, organ rendering, and a virtual ``radiologist'' avatar delivering reformatted explanations (Fig. \ref{fig:teaser}). This multi-modal approach not only aims to improve patient understanding but also closely simulates the experience of a one-on-one consultation with a healthcare provider.

\textbf{Evaluation and Significance.}
\textbf{Crucially, our work includes an early evaluation of the system's usefulness in a clinical context.} We conducted a proof-of-concept study involving six practicing radiologists to assess the potential impact and practicality of our approach. This evaluation provides valuable insights into the system's strengths, limitations, and potential for real-world application in patient-centered radiology.

%% file: content/02-relatedwork.tex
\section{Related Works}

\textbf{Radiology Report Understanding.}
A study by \citet{miller2013enhancing} showed that direct radiologist-patient communications enhanced patients' experience and could improve patients' care.
However, this approach would further strain radiologists' workload, exacerbating the existing disparity between growing medical imaging service demands and limited radiologist capacity \citep{heiser2019new}.
With the growing adoption of patient portals providing direct access to radiology reports, many approaches have been explored to enhance patients' comprehension of their medical imaging findings.
For example, \citet{cook2017patients} provided additional explanations of medical terminologies to patients, which was found able to help them understand their reports.  
\citet{lourenco2020optimizing} suggested using structured reports to improve the clarity and ease the information extraction for patient audience. 

\begin{figure*}[!t]
  \includegraphics[width=\linewidth]{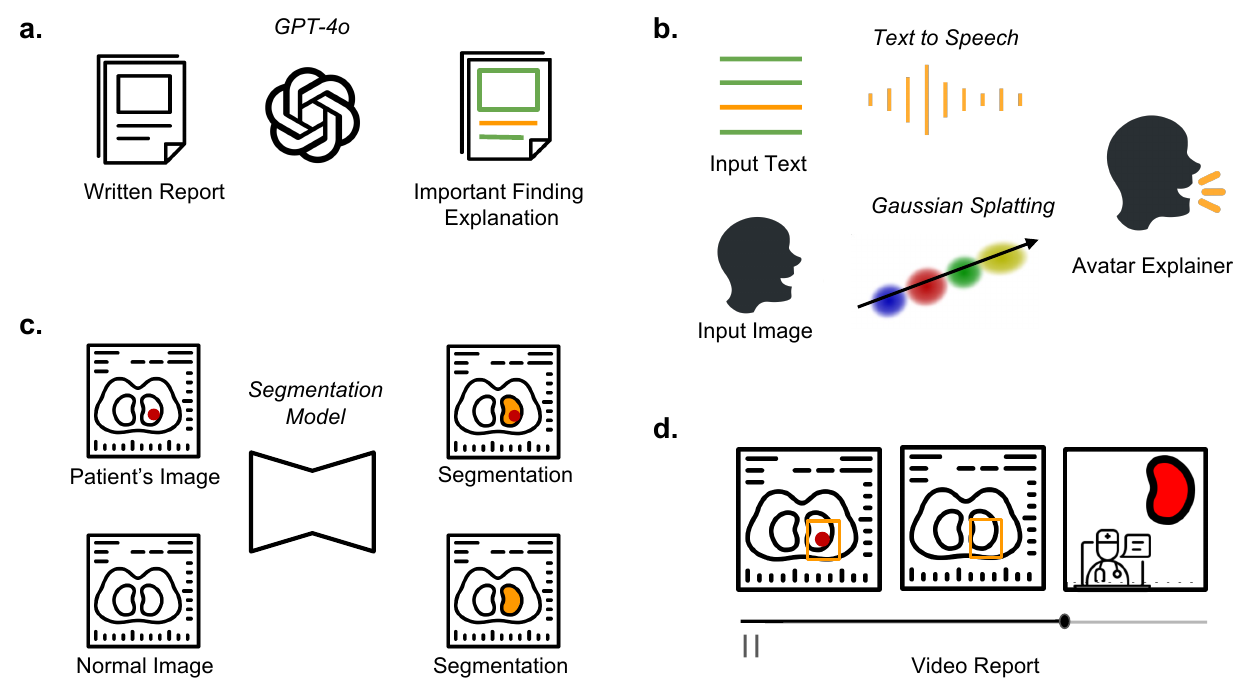}
  \caption{\textbf{Illustration of the ReXplain pipeline.} \textbf{a}. GPT-4o is first used to generate lay-language explanation from the text reports. It is also used to map the findings to the corresponding organs; \textbf{b}. Avatar generation based on text to speech generation and Gaussian Splatting, which simulates a one-on-one communication interface; \textbf{c}. A universal CT organ segmentation model is used to highlight the anatomy of interest according to the organ extracted from the previous steps; \textbf{d}. The final video report combining the key elements generated from the previous steps.}
  \label{fig:framework}
\end{figure*}

\textbf{AI in Radiology Communication.}
Recent advances in artificial intelligence (AI) have shown promise in bridging the communication divide between radiologists and patients by providing patients with direct access to their reports.
Particularly, the advent of Large Language Models (LLMs) has opened new possibilities in this field, with proven ability to simplify the radiology reports while largely maintaining factual correctness and completeness \citep{lyu2023translating,doshi2023utilizing,jeblick2024chatgpt} as well as potential in improving patients engagement \cite{elkassem2023potential}.
The representative LLM, GPT-4 \citep{achiam2023gpt}, was also reported to be able to identify and generate patient instructions for actionable incidental radiology findings \citep{woo2024evaluation}.
These developments offer new possibilities for tackling the limitations of current laborious approaches and could allow patients to better process and understand their own medical data \citep{topol2019high,rajpurkar2023current}.
Nonetheless, the current AI approaches focus on explaining medical text, and few has extended the explanation to multimodal data with medical images.

\textbf{Radiology Video Reports.}
Access to their radiology images can enhance patients' understanding of the reports, as shown in the prior study by \citet{halaska2019patient}. 
However, connecting text descriptions to findings on radiology images remains challenging for patients without medical training, limiting their comprehension of highly specialized medical information. 
A promising solution to this is generating radiology video reports to highlight important findings and connect them with the written findings \citep{balkman2016audio,neto2019evaluation,recht2022video}. Notably, \citet{recht2022video} found that video reports made by radiologists, featuring explanations and visual guidance, were preferred by patients over written reports alone. Nevertheless, these approaches require additional effort from radiologists, which may be impractical given their already heavy workload, especially for volumetric images like CT and MRI.

%% file: content/02-design.tex
\section{Designing ReXplain}

\paragraph{Sketching Video Reports Concepts.} In typical clinical practice, radiologists guide patients through the findings written on the report and show them with the corresponding images simultaneously. To analyze this interaction, we simulated a patient consultation where a radiologist recorded his explanations based on a retrospective CT scan paired with report from the CT-RATE \citep{hamamci2024developing} dataset.
Noticeably, there were several steps that help improve the understanding, which involved typically (1) explaining the findings to the patient with plain language; meanwhile (2) pointing at the abnormalities on the image; and (3) supplementing necessary explanations on how the image should look if there was no abnormalities.

Building on these observations, we conceptualized an AI-driven approach to recreate this educational experience through automated video reports. Our design focuses on four technical challenges in generating patient-friendly video explanations: 
(1) translation of medical terminology and complex findings into clear, patient-friendly descriptions that preserve clinical accuracy while enhancing comprehension;
(2) automated identification and visual annotation of relevant findings on medical images to guide patient attention to specific areas of interest;
(3) integration of normal reference images alongside abnormal findings to help patients understand the nature and significance of their condition through direct comparison;
(4) seamless integration of visual elements, annotations, and explanatory text through an intuitive interface that maintains a coherent narrative flow.
The technical implementation of these components is detailed in subsequent sections.

\paragraph{Interpreting Radiology Reports with Lay-language.}
The first goal is to interpret the radiology reports to be easy-to-understand by non-experts.
Here, we leveraged the large language model, GPT-4o \citep{achiam2023gpt}, which has been reported capable of lowering the reading level of professional medical documents.
First, we required GPT-4o to extract phrases that describe positive findings, which are supposed to be information of the most interest for the patients.
Then, we prompted GPT-4o to imitate radiologists's explanation and generate three types of messages in lay language: (1) an explanation of what findings are detected; (2) a description of how the findings appear on a CT scan; (3) a description of how the CT scan should appear without the abnormal findings.
This procedure mimics the radiologist-patient communication of explaining the radiological findings.
In this way, we managed to rephrase the specialist radiology reports to have higher readability with structured presentation order, which is prepared to be simultaneously presented with images displayed later.
The prompts we used can be found in the Appendix.

\paragraph{Connecting Image Regions with Reports.}
To imitate radiologists' explanation for CT images, we would need to localize the regions of interest (ROIs) for an intuitive visualization of the report findings.
Currently, there are few AI models that can ground the lesions with free-text radiology report, we thus relaxed the requirement from localizing the findings to localizing the anatomic structures that contain the findings.
To achieve this, we first prompted GPT-4o to match the extracted findings to one of the organs from a finite set of 201 human organs (a list of the organs can be found in the Appendix).
Then, to achieve comprehensive localization across a wide range of anatomical structures, we utilized the recently developed AI model, the Segment Anything in medical images via Text model (SAT) \citep{zhao2023one}.
Specifically, SAT leverages the capability of natural language prompts to effectively segment anatomical structures from 3D medical volumes. This knowledge-enhanced segmentation model had been trained on large-scale CT data for whole-body anatomy segmentation.
After obtaining the segmented organ, we utilized PyTorch3D \citep{ravi2020pytorch3d} to generate 3D rendering visualizations to further provide an overview of the anatomical structure.
By using SAT segmentation masks and the matching of report descriptions to organs, we constructed a mechanism to connect the text reports to regions on the images, which will be further utilized to construct the video presentation.

\paragraph{Comparing with a Healthy Individual.}
To supplement the explanation with reference to normal images, we retrieved an CT scan of a healthy individual from the CT-RATE dataset and conducted organ segmentation following the previous procedure.
During the video generation, we register this normal scan with each new input image and morph the segmentation masks correspondingly. 
The rigid registration model by SimpleElastix \citep{marstal2016simpleelastix} was used for registration.

\paragraph{Generating Avatar Explainer.}
To facilitate more effective dissemination and reception of complex medical information, we enhanced our instructional videos with an AI-generated virtual explainer. This avatar would simulate a medical practitioner, expected to improve patient comprehension and promote greater acceptance of the content.
Building upon the translated report from previous steps, this step leveraged a text-to-video avatar generation model.
Particularly, we took advantage of the commercially ready avatar generation pipeline, Tavus \citep{tavus2024advanced,tavus2024phoenix}, which consists of a series of AI techniques, including text to speech generation \citep{tan2021survey}, 3D head and shoulders reconstruction, audio to facial animation generation, and avatar video rendering utilizing 3D Gaussian Splatting \citep{kerbl20233d}.
By employing 3D Gaussian Splatting, the system efficiently handles complex visual features such as skin texture, hair, and subtle facial nuances. 
This resulted in an virtual explainer that enhances the engageability of the video reports with a simulated one-on-one conversation.

\begin{figure*}[!t]
\centering
  \includegraphics[width=\linewidth]{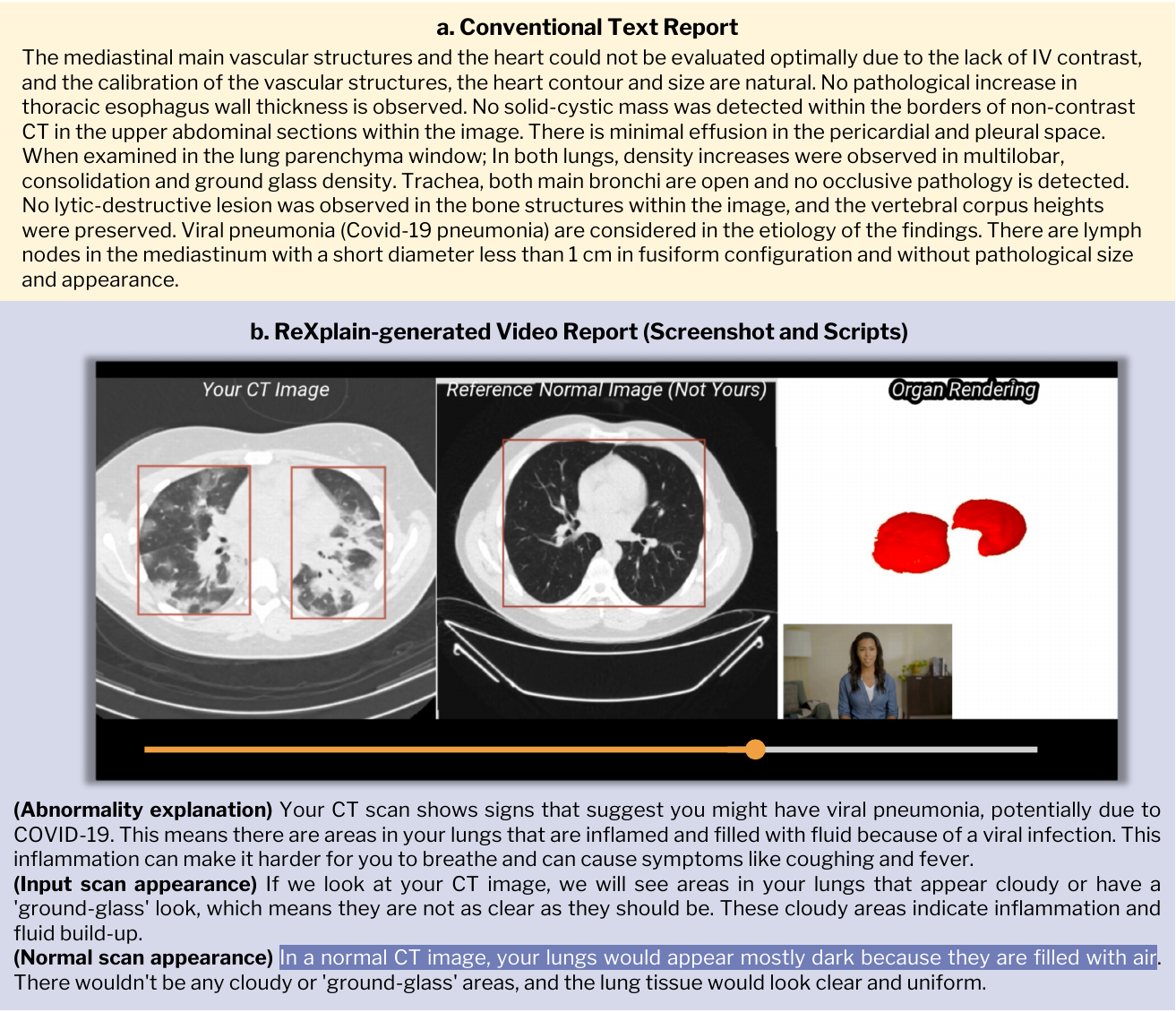}
  \caption{\textbf{Comparison between a conventional written report and an illustration of a video report generated by ReXplain.} \textbf{a}. The original report describes both negative and positive findings. \textbf{b}. The video report highlights positive findings with grounded image regions, emphasizes comparison with normal images, renders the holistic organ structure, and explains the findings using lay language with avatar explainer. A sample video report can be found in the supplementary.}
  \label{fig:report_show}
\end{figure*}

\begin{table*}[!t]
\centering
\begin{tabular}{@{}p{0.02\textwidth}p{0.8\textwidth}c@{}}
\toprule
\multicolumn{3}{c}{\textbf{Video Reports Evaluation Questionnaire}} \\
\addlinespace
\textbf{\#}& \multicolumn{2}{l}{\textbf{Please rate 5-1 (5: strongly agree, 1: strongly disagree) for the following questions:}}\\
\midrule
\addlinespace
1 & The video correctly identifies the important findings in the CT. & $\square$ \\
\addlinespace
2 & The video sufficiently reviews the findings with the patient. & $\square$ \\
\addlinespace
3 & The video correctly localizes the findings to the appropriate organ. & $\square$ \\
\addlinespace
4 & The video explains the findings in a way that can be understood by a patient (assuming an eighth-grade reading level). & $\square$ \\
\addlinespace
5 & I am comfortable walking my patients through this video to help them understand their findings. & $\square$ \\
\addlinespace
6 & I am comfortable showing the videos to my patients without my supervision. & $\square$ \\
\addlinespace
7 & This video comes across as conversational with a patient. & $\square$ \\
\addlinespace
8 & The explanation of the report is easy to understand. & $\square$ \\
\addlinespace
9 & The comparison with normal CT scan improves understanding of the condition. & $\square$ \\
\addlinespace
10 & Connecting the explanation with the image helps improve understanding. & $\square$ \\
\addlinespace
11 & The rendering of the organ helps understand the 3D structure. & $\square$ \\
\addlinespace
12 & The avatar is natural and conversational. & $\square$ \\
\bottomrule
\end{tabular}
\caption{\textbf{Video Reports Evaluation Questionnaire.}}
\label{tab:video_evaluation}
\end{table*}

\begin{figure*}[!t]
\centering
  \includegraphics[width=.9\linewidth]{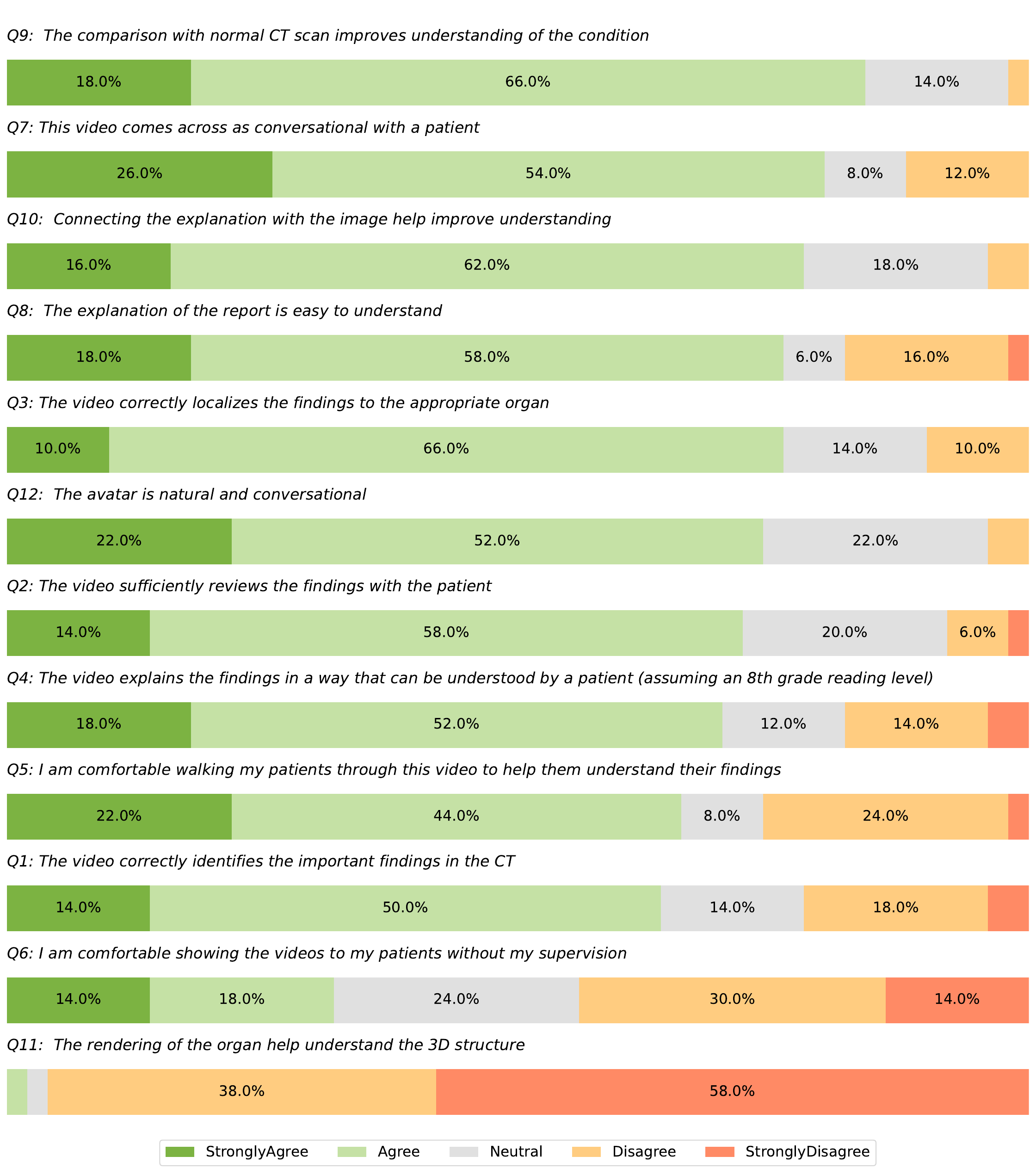}
\caption{\textbf{Round 1 user study results.} Questions are sorted by the descending order of the percentage of the combination of ``strongly agree'' and ``agree''. Neutral = ``neither agree or disagree''.}
\label{fig:user_study}
\end{figure*}

%% file: content/03-output.tex
\section{Generating Patient-friendly Video Reports}

\paragraph{Integrating Key Elements into Video Reports.}
The former pipeline composes ReXplain (Fig. \ref{fig:framework}), based on which we designed a comprehensive video logic to effectively convey radiology image findings to patients, in a way mimics the radiologists.
As mentioned, the finding is presented using easily understandable lay language, ensuring that the information is accessible to those without medical backgrounds. 
We adjusted the frames per second of the CT displaying to synchronize it with the avatar explanation.
Meanwhile, we also added the display of the organ rendering to illustrate the global view of the anatomy of interest.
The CT windows were set to change with the findings, e.g., lung window would be used to display findings on the lungs.
The final video represents an integrated presentation of visual and auditory information, which will help reinforce the explanation of the radiology reports.

\paragraph{Exemplifying a Typical Video Report.}
Fig. \ref{fig:report_show} compares a conventional text report and the ReXplain-generated video report with a case containing COVID-19 pneumonia.
As can be observed, a conventional CT report contains many negative findings describing that the investigated regions are normal.
On the contrary, the video report generated by ReXplain focuses on the positive findings, where the explanation of the COVID-19 findings are illustrated in Fig. \ref{fig:report_show}b.
Specifically, the interface contains three main panels. The first panel displays the patient's imaging with bounding boxes highlighting the infected lung areas. The second panel displays a normal image for a comparison. The third panel shows rotating 3D rendering of the organ, \textit{i.e.}, the lungs, as well as the visual avatar in the bottom left. The avatar narrates the translated report as the imaging is displayed.

A video report contains three phases. In the first phrase, the patient's radiology image is presented with a bounding box highlighting the region of interest. The avatar introduces the finding, explains what it is and how it may affect the patient. 
The screen displays the region of interest by scrolling the CT scans in the axial view. In the second part, the CT scan scrolls again, with the avatar explaining how the finding appears on the image. In the third phase, the input image and a comparison normal image are displayed together with bounding boxes to highlight their differences, and the avatar describes how a normal image should appear.
When multiple positive findings are observed, we start with the most clinically significant findings, which are often the most pressing patient concerns, and then proceed to less severe findings.

\section{Technical Performance}

We conducted a manual evaluation on the output of GPT-4, focusing on 1) whether the extracted phrases were describing abnormal findings; 2) whether the phrases describing abnormalities were correctly extracted; and 3) whether the organs containing the abnormalities were correctly connected. We randomly selected the findings section in 50 reports from CT-RATE, of which 204 sentences describes abnormalities.
Using the designed prompt, GPT-4 extracted 174 sentences, all of which describe abnormalities. This means that GPT-4 achieves 100\% precision and 85\% recall on abnormality extraction. In addition, all extracted phrases were correctly matched to the corresponding organs, yielding a 100\% accuracy for finding-region connection. On the other hand, essentially, for 23 reports that contained less than or equal to 2 abnormality descriptions, GPT-4 achieved 100\% precision and 100\% recall on abnormality extraction, and 100\% accuracy on finding-region connection.
These results demonstrate the technical feasibility of the proposed ReXplain pipeline.
In the following user studies, we used samples of which the reports contains at most three abnormal findings.
The segmentation model, SAT, was reported to achieve 84.24\% Dice score in whole-body segmentation \cite{zhao2023one}, which could satisfy the requirement of highlighting organs for ReXplain.

%% file: content/04-user.tex
\section{Eliciting User Feedback}

\paragraph{Materials.}
The CT images used for this study were sampled from CT-RATE~\citep{hamamci2024foundation}, a publicly available dataset of non-contrast 3D chest CT volumes paired with radiology text reports.
We standardized all CT volumes to a uniform voxel spacing of $1\times1\times3$ mm to meet the input requirement by SAT.
As the data from CT-RATE could partially cover the abdomen region, we amended the prompt to match the report findings to 201 specific anatomical regions covering both the thorax and abdomen. The regions are listed in the appendix.

\paragraph{Ethics Statement}
This work was based on publicly available data with no identifiable personal information.
The user study was conducted exclusively with the participation of the authors, all of whom are investigators on this project.
No external participants were involved, and no personally identifiable information was collected or used. As the current research involved only the investigators themselves, it did not constitute human subjects research requiring Institutional Review Board (IRB) approval. 

\subsection{Round 1}
\paragraph{User Study Design}

We conducted a pilot user study to evaluate the effectiveness of the video reports.
Specifically, we selected ten samples from different patients. The resulting ten video reports are at most three minutes long, explaining one or two abnormal findings.
Then, both the video reports and the original text reports were provided to the radiologist team (round 1 involves four radiologists with 6-10 years of practice, and one with 11-15 years of practice).
Each pair of reports is associated with a survey which would be completed after the reports being reviewed and compared.

We prepared 12 questions for each video, which can be found in Table \ref{tab:video_evaluation}.
Specifically, questions \#1-6 evaluated the correctness of the information delivered and its suitability for patient communication, while questions \#7-12 investigated the helpfulness of each element of video.
For each question, we rated our level of agreement from 5 to 1, representing “strongly agree,” “agree,” “neither agree nor disagree,” “disagree,” and “strongly disagree,” respectively. In addition to direct scoring, we also provided open-ended questions for more flexible feedback, inviting radiologist team members to offer written comments on “which parts of the video were most useful or valuable” and “how would you improve this video?” This approach allowed us to gain a deeper understanding of the collective expert opinions.

\paragraph{ReXplain is Useful in Enhancing Understanding.}
Based on ten videos and feedback from five radiologists, we collected 50 data points for each of the questions.
We plotted the distribution of the scores in Fig. \ref{fig:user_study} and sorted the questions in descending order of the percentage of the combination of ``strongly agree'' and ``agree''.

Specifically, there were at least 64\% positive feedback (``agree'' or ``strongly agree'') on questions \#1-4, indicating that the information delivered by the videos are largely correct and easy-to-understand.
These results reveal that the video reports can be potentially used to convey the radiology findings to patients without medical knowledge.

Moreover, questions \#8-10 and \#12 received at least 74\% positive feedback, indicating the useful components in the videos and answered how the video reports improved patient-friendly radiology.
The written comments also showed that \textit{``showing comparison with normal image''} and \textit{``the layperson explanation of the report''} were the two most frequent positive feedback.
These results showed that most elements in the video reports, \textit{i.e.}, the translated reports with lay language, the connection between image regions and reported findings, the comparison to the normal images, and the use of the avatar explainer, could effectively assist in improving the understanding and engagement of audience.

\paragraph{ReXplain Shows Potential in Clinical Practice.}
In clinical practice, limited time and heavy workload often prevents radiologists from directly communicating with patients \citep{kemp2017patient}.
The study by \citet{recht2022video} showed that hand-crafted video reports by radiologists were helpful in improving patients' understanding.
However, this approach offered limited information with increased workload (each video had 55 ± 30 seconds of duration and was generated in 238 ± 141 seconds).
In contrary, ReXplain functions end-to-end without requiring extra time or effort from radiologists, offering enhanced scalability in practice.
Furthermore, the explanation could be tailored to provide flexibility for customization through appropriate prompting.
We observed that 66\% of responses were optimistic about walking patients through the videos (Question \#5), which indicates that the generated video reports effectively conveyed the information we intended for patients to understand.
This result further demonstrated that ReXplain holds promise in improving patients understanding and experience with lighter workload on radiologists.

\subsection{Round 2}

\paragraph{User Study Design.}
Based on feedback from Round 1, we conducted a second round of user feedback to gain deeper insights into the proposed pipeline. We used a structured survey with open-ended questions that explored ReXplain's limitations, potential improvements, and clinical applications, focusing on exploring features that might discourage patient sharing, addressing safety considerations, and discussing the usefulness of videos without pathology highlighting. \textcolor{black}{Four radiologists participated in this round, including two with 6-10 years of practice and two with 11-15 years of experience. Three of these radiologists had previously completed the Round 1 survey.}

\paragraph{Current Limitations and Potential Improvements.} The current organ-level segmentation approach was identified as a key limitation, with radiologists emphasizing the need for more precise pathology highlighting and visualization. 
A significant portion of written feedback from the radiology team suggested that highlighting entire organs may have limited effectiveness in improving understanding, particularly for small findings such as lung nodules.
This probably led to predominantly negative feedback in Round 1 regarding the usefulness of organ rendering for patient comprehension (question \#11) and that only 32\% of responses (question \#6) indicated acceptance of allowing patients to view the videos without supervision.
We deem that the limitation is majorly caused by the current state of medical image segmentation models, as few support free-text-based grounding of lesions.
While we believe that this challenge can be addressed with advancements in text-prompted lesion segmentation, such as BiomedParse \citep{zhao2024biomedparse}. 
Another limitation pointed is that the continuous video scrolling made it challenging for viewers to focus on specific findings, suggesting a need for strategic pausing at key points.
Also, this can be addressed by improving radiology image segmentation to the lesion level, which would enable key slice display focusing on the findings. The feedback also highlighted the importance of avoiding misleading severity assessments and integrating trusted radiology information sources. 
Improvement was also suggested to implement variable explanation complexity levels and including clear directives for patients to consult healthcare providers.

\paragraph{Potential Clinical Usage.}
Building on earlier discussions, ReXplain is envisioned as a pre-consultation educational tool for radiologists.
It is suggested to provide ReXplain-generated video reports to patients \textit{``probably immediately before the physician sees them.''}, followed by \textit{``in-person review and question-answer session''}, to secure its usage.
Moreover, one of the collected suggestions is \textit{``these available with every imaging study and look for improvements in patient engagement and follow up, and decreased burden on the referring provider (who ordered the study and typically would have to explain/discuss with the patient)''}.
While it was emphasized that precise lesion identification would significantly enhance clinical value, \textcolor{black}{three out of four} respondents maintained that the videos retain utility even without specific pathology highlighting.
These insights reinforce our initial findings while providing concrete directions for technical improvements and safety considerations in clinical deployment.

\paragraph{Outlook.}
This proof-of-concept study aimed to to figure out whether and how the AI-driven pipeline, ReXplain, could help translate the conventional radiology reports into patient-friendly video reports. Nevertheless, our user study was conducted by radiologists, of whom the feedback may not precisely reflect the experience of real patients. The current work shows that the video reports overall were considered useful for enhancing patients' understanding, and this finding motivates us to conduct future works involving patients without medical knowledge to further investigate the usefulness of the AI-generated video reports.

%% file: content/05-discussion.tex
\section{Conclusion}

In this paper, we present ReXplain, an innovative system that integrates state-of-the-art modules to produce patient-friendly radiology video reports. Our end-to-end pipeline combines advanced AI capabilities with a practical interface design to address a critical healthcare communication challenge: helping patients better understand their diagnostic results. A proof-of-concept study with six board-certified radiologists yielded encouraging feedback on the utility of these AI-generated video reports. Through the development and initial assessment of ReXplain, we lay the groundwork for AI-assisted, patient-centered radiology reporting.